# Open Source Dataset and Machine Learning Techniques for Automatic Recognition of Historical Graffiti

Nikita Gordienko[1], Peng Gang[2], Yuri Gordienko*[1], Wei Zeng[2], Oleg Alienin[1], Oleksandr Rokovyi[1], and Sergii Stirenko[1]

[1]National Technical University of Ukraine "Igor Sikorsky Kyiv Polytechnic Institute", Kyiv, Ukraine
[2]School of Information Science and Technology, Huizhou University, Huizhou, China
*yuri.gordiienko@gmail.com

**Abstract.** Machine learning techniques are presented for automatic recognition of the historical letters (XI-XVIII centuries) carved on the stoned walls of St.Sophia cathedral in Kyiv (Ukraine). A new image dataset of these carved Glagolitic and Cyrillic letters (CGCL) was assembled and pre-processed for recognition and prediction by machine learning methods. The dataset consists of more than 4000 images for 34 types of letters. The explanatory data analysis of CGCL and notMNIST datasets shown that the carved letters can hardly be differentiated by dimensionality reduction methods, for example, by t-distributed stochastic neighbor embedding (tSNE) due to the worse letter representation by stone carving in comparison to hand writing. The multinomial logistic regression (MLR) and a 2D convolutional neural network (CNN) models were applied. The MLR model demonstrated the area under curve (AUC) values for receiver operating characteristic (ROC) are not lower than 0.92 and 0.60 for notMNIST and CGCL, respectively. The CNN model gave AUC values close to 0.99 for both notMNIST and CGCL (despite the much smaller size and quality of CGCL in comparison to notMNIST) under condition of the high lossy data augmentation. CGCL dataset was published to be available for the data science community as an open source resource.

## 1 Introduction

Various writing systems have been created by humankind, and they evolved based on the available writing tools and carriers. The term graffiti relates to any writing found on the walls of ancient buildings, and now the word includes any graphics applied to surfaces (usually in the context of vandalism) [1]. But graffiti are very powerful source of historical knowledge, for example, the only known source of the Safaitic language is graffiti inscriptions on the surface of rocks in southern Syria, eastern Jordan and northern Saudi Arabia [2]. In addition to these well-known facts, the most interesting and original examples of the Eastern Slavic visual texts are represented by the medieval graffiti that can be found in St. Sophia Cathedral of Kyiv (Ukraine) (Fig.1) [3].They are written in two alphabets, Glagolitic and Cyrillic, and vary by the letter style, arrangement and layout [4,5].The various interpretations of these graffiti were suggested by scholars as to their date, language, authorship, genuineness, and meaning [6,7]. Some of them were based on the various image processing techniques including pattern recognition, optical character recognition, etc.

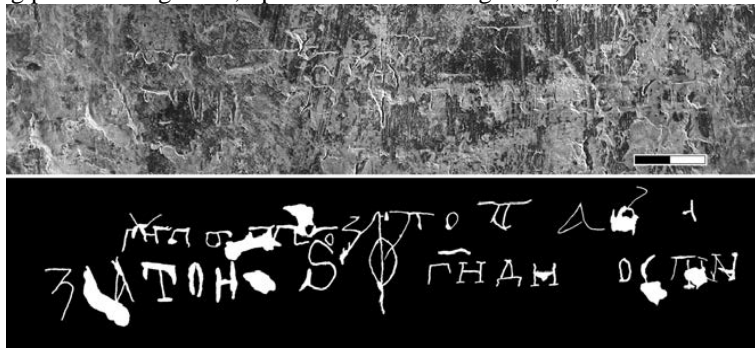

**Fig. 1.** Example of the original image for graffito #1 (c. 1022) (a) and preprocessed glyphs (b) from the medieval graffiti in St. Sophia Cathedral of Kyiv (Ukraine) [3].

The main aim of this paper is to apply some machine learning techniques for automatic recognition of the historical graffiti, namely letters (XI-XVIII centuries) carved on the stoned walls of St.Sophia cathedral in Kyiv (Ukraine) and estimate their efficiency in the view of the complex geometry, barely discernible shape, and low statistical representativeness (small dataset problem). The section 2.*State of the Art* contains the short characterization of the basic terms and parameters of the methods used. The section 3.*Datasets and Models* includes description of the datasets used and methods applied for their characterization. The section 4.*Explanatory Data Analysis* gives results of the initial analysis of preprocessed graffiti images. The section 5.*Machine Learning for Automatic Recognition of Graffiti* contains results of machine learning approaches to the problem. The section 6.*Discussion and Future Work* is dedicated to discussion of the results obtained and lessons learned.

## 2    State of the Art

The current and previous works on recognition of handwriting were mainly targeted to pen, pencil, stilus, or finger writing. The high values of recognition accuracy (>99%) were demonstrated on the MNIST dataset [8] of handwritten digits by a convolutional neural networks [9]. But stone carved handwriting has usually much worse quality and shabby state to provide the similar values of accuracy. At the moment, most work on character recognition has concentrated on pen-on-paper like systems [10]. In all cases the methods were based on the significant preprocessing actions without which accuracy falls significantly. And this is especially important for analy-sis and recognition of the carved letters like historical graffiti. Usually, the preprocessing requires a priori knowledge about entire glyph, but the Glagolitic and Cyrillic glyph datasets are not available at the moment as open source databases except for some cases of their publications [3,4]. The recent progress of computer vision and machine learning methods allows to apply some of them to improve the current recognition, identification, localization, semantic segmentation, and interpretation of such historical graffiti of various origin from different regions and cultures, including Europe (ancient Ukrainian graffiti from Kyivan Rus) [3], Middle East and Africa (Safaitic graffiti) [2], Asia (Chinese hieroglyphs) [11], etc. Moreover, the progress of diverse mediums in the recent decades determined the need for many more alphabets and methods of their recognition for different use cases, such as controlling computers using touchpads, mouse gestures or eye tracking cameras. It is especially important topic for elderly care applications [12] on the basis of the newly available information and communication technologies based on multimodal interaction through human-computer interfaces like wearable computing, brain-computing interfaces [13], etc.

## 3 Datasets and Models

Currently, more than 7000 graffiti of St. Sophia Cathedral of Kyiv are detected, studied, preprocessed, and classified (Fig.1) [14-16]. The unique corpus of epigraphic monuments of St. Sophia of Kyiv belongs to the oldest inscriptions, which are the most valuable and reliable source to determine the time of construction of the main temple of Kyivan Rus. For example, they contain the cathedral inscriptions-graffiti dated back to 1018–1022, which reliably confirmed the foundation of the St. Sophia Cathedral in 1011.A new image dataset of these carved Glagolitic and Cyrillic letters (CGCL) from graffiti of St. Sophia Cathedral of Kyiv was assembled and preprocessed to provide glyphs (Fig.1a) for recognition and prediction by multinomial logistic regression and deep neural network [17]. At the moment the whole dataset consists of more than 4000 images for 34types of letters (classes), but it is permanently enlarged by the fresh contributions.

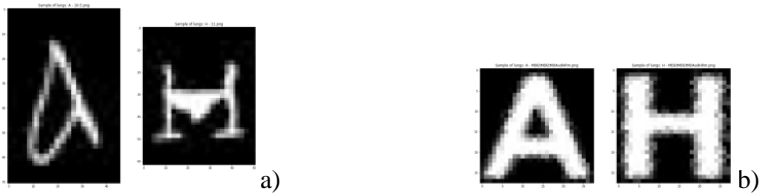

a)                                                                    b)

**Fig. 2.** Examples of glyphs obtained from: CGCL dataset from graffiti of St. Sophia Cathedral of Kyiv (a) and from notMNIST dataset (b).

The second dataset, notMNIST, contains some publicly available fonts for 10 classes (letters A-J) taken from different fonts and extracted glyphs from them to make a dataset similar to MNIST [8]. notMNIST dataset consists of small (cleaned) part, about 19k instances (Fig.1b), and large (unclean) dataset, 500k instances [18].It was used for comparison of the results obtained with GCCL dataset.

## 4      Explanatory Data Analysis

The well-known t-distributed stochastic neighbor embedding (t-SNE) technique was applied [19,20]. It allowed us to embed high-dimensional glyph image data into a 3D space, which can then be visualized in a scatter plot (Fig.3).The cluster of glyphs of the carved graffiti from CGCL dataset is more scattered (Fig.3a) than the cluster of glyphs from notMNIST dataset(Fig.3b) (note the difference of >30 times for scales of these plots).

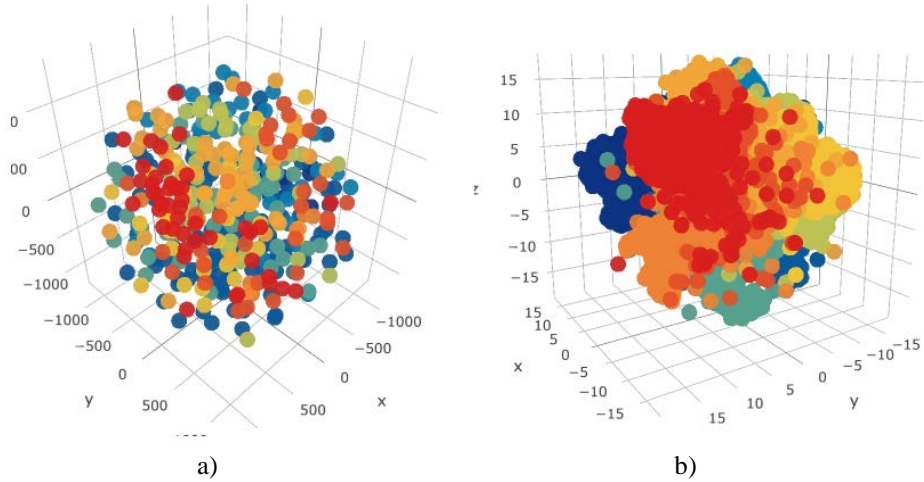

a)                                                    b)

**Fig. 3.** Results of tSNE analysis for 10 letters from A (red) to H (blue) glyphs (Fig.2) from: CGCL dataset (a) and notMNIST dataset (b).The similar letters are modeled by nearby points and dissimilar ones are mapped to distant points. The same color corresponds to the same class (type of letter from A to H).

The explanatory data analysis of CGCL and notMNIST datasets shown that the carved letters can hardly be differentiated by dimensionality reduction methods, for example, by t-distributed stochastic neighbor embedding (tSNE) due to the worse letter representation by stone carving in comparison to hand writing.

For the better representation of the distances between different images the cluster analysis of differences by calculation of pairwise image distances was performed for subsets of the original CGCL dataset and notMNIST datasets that contained glyphs of A and H letters only. Then the clustered distance map was constructed for CGCL dataset (Fig.4a) and notMNIST (Fig.4b)datasets. The crucial difference of the visual quality of glyphs from CGCL and notMNIST datasets consists in the more pronounced clustering in two distinctive sets (denoted by separate blue and red parts of the legend ribbons) and more darker regions inside map (the darker region means the lower distance between letters) for notMNIST dataset (Fig.4b). Even from the first look these maps are very different and have the clear understanding about correlation among the glyphs in both datasets.

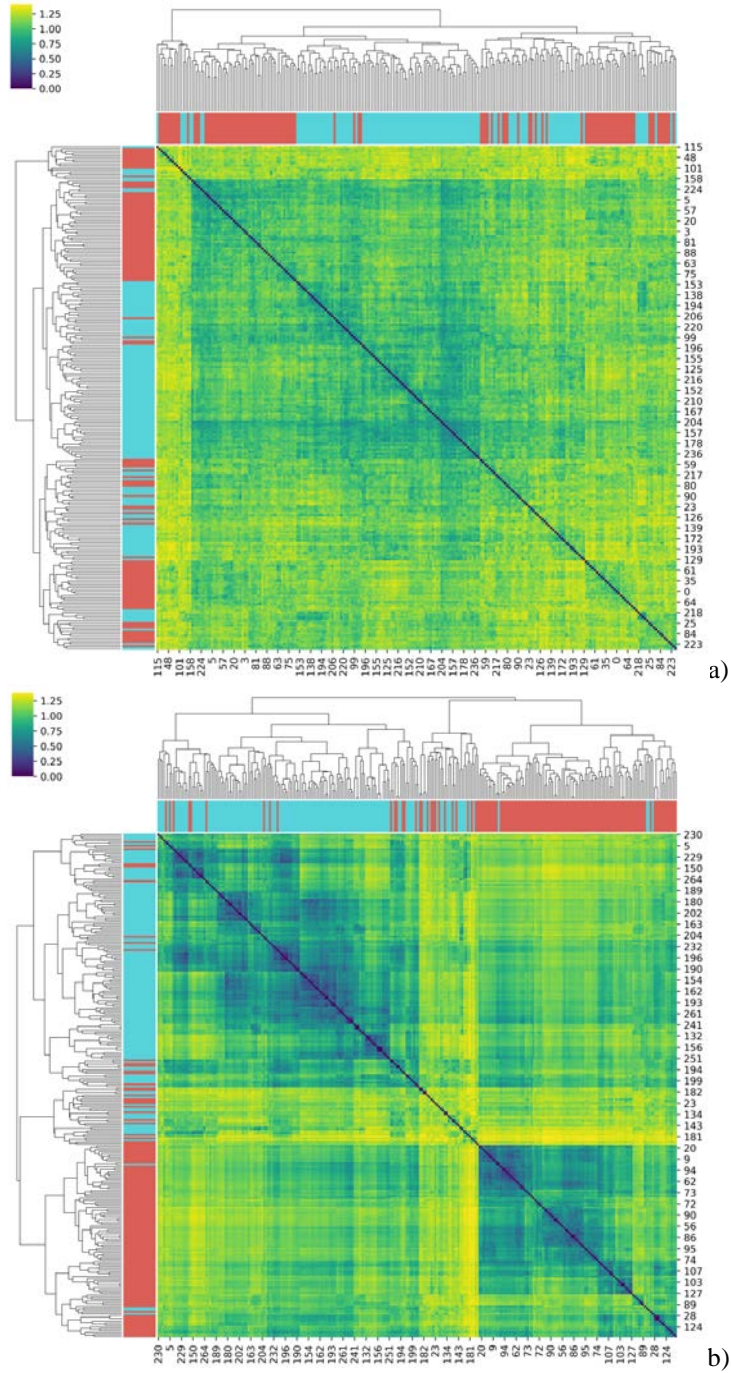

**Fig. 4.** The clustered distance maps for distances between A (red) and H (blue) letters in CGCL (a) and notMNIST (b) datasets.

## 5    Machine Learning for Automatic Recognition of Graffiti

### 5.1    Multinomial Logistic Regression

To estimate the possibility to predict the letters by glyphs the multinomial logistic regression (MLR) was applied for subsets with 10 classes of letters (Fig.5). The MLR model demonstrated that the area under curve (AUC) values for receiver operating characteristic (ROC) for separate letters were not lower than 0.92 and 0.60 for notM-NIST and CGCL, respectively, and the averaged AUC values were 0.99 and 0.82 for notMNIST and CGCL, respectively.

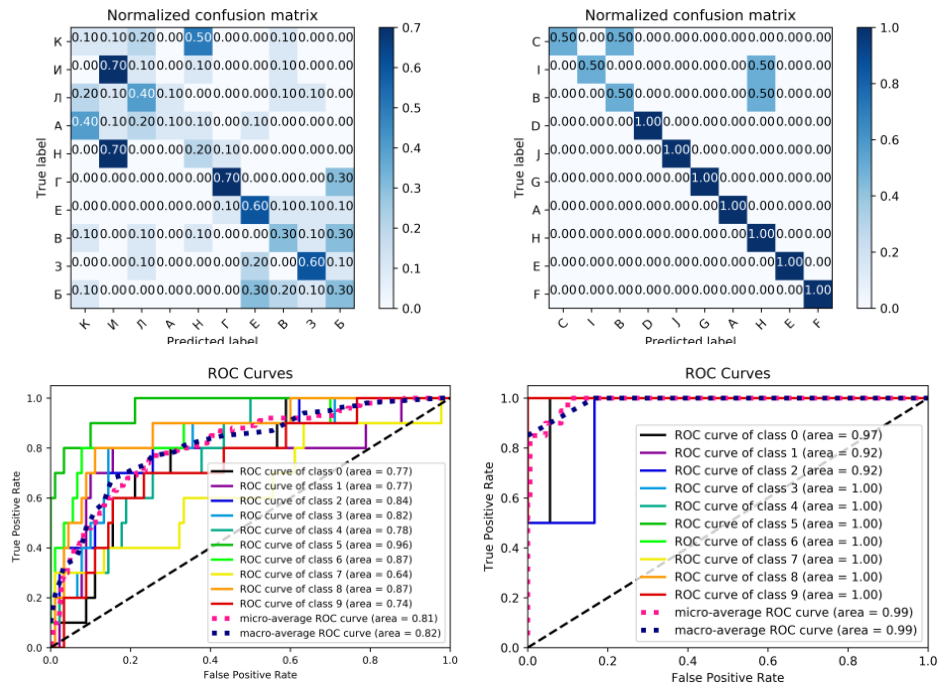

**Fig. 5.** Confusion matrixes and ROC-curves for CGCL (left) and notMNIST (right) datasets.

### 5.2    Convolutional Neural Network

2D convolutional neural network (CNN) was applied to check the feasibility of application of neural networks for the small dataset like CGCL in comparison to notM-NIST dataset to recognize two letters A and H from their glyphs. The CNN had pyramid like architecture with 5 convolutional/max-pooling layers and 205 217 trainable parameters, rectified linear unit (ReLU) activation functions, a binary cross-entropy as a loss function, and RMSProp (Root Mean Square Propagation) as an optimizer with a learning rate of $10^{-4}$. In Fig.6 accuracy and loss results of training and validation attempts are shown for subsets of the original CGCL (Fig.6, left) and notMNIST (Fig.6, right) datasets that contained glyphs of A and H letters only.

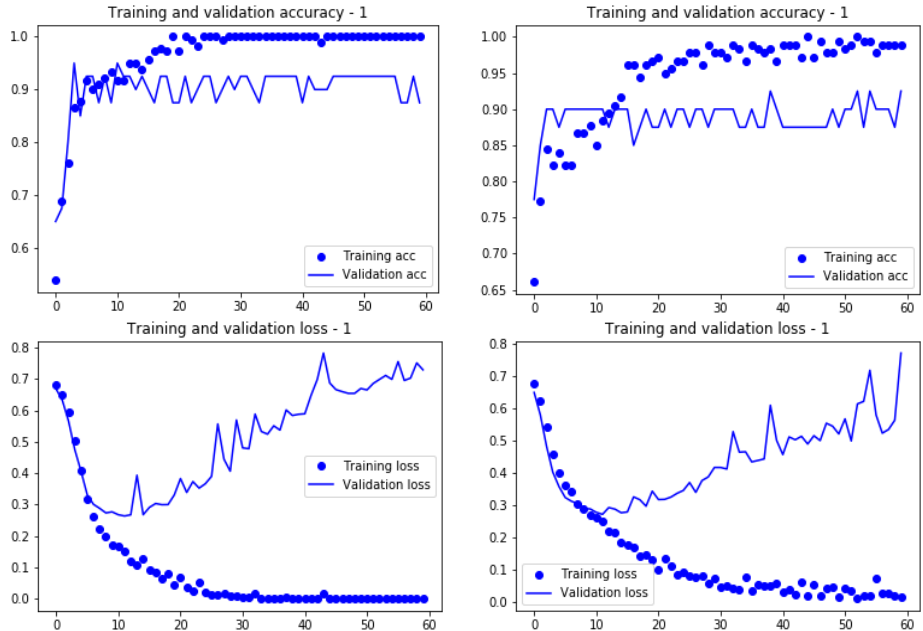

**Fig. 6.** The accuracy and loss for the original CGCL (left) and notMNIST (right) datasets.

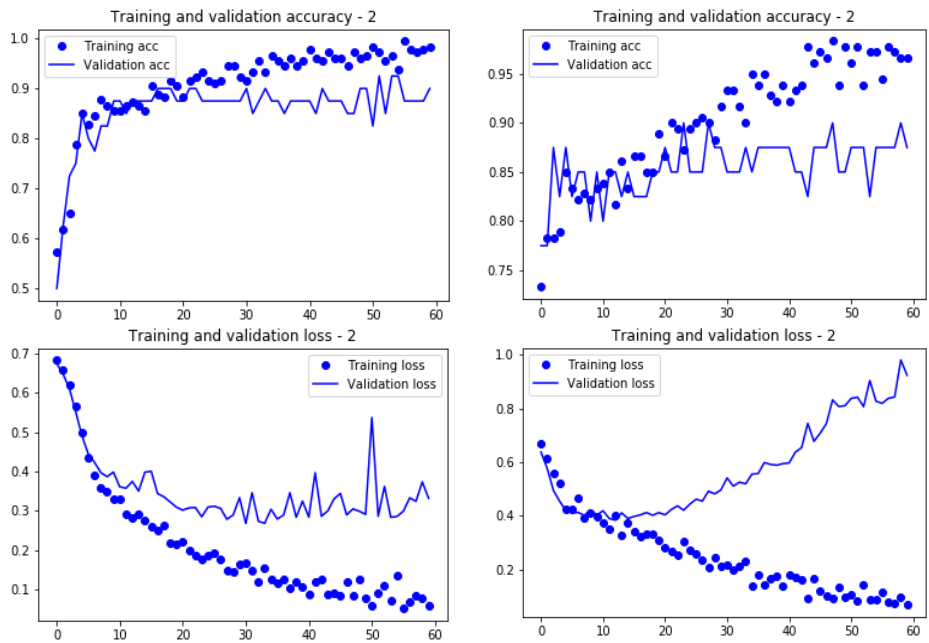

**Fig. 7.** The accuracy and loss for CGCL (left) and notMNIST (right) with lossless data augmentation.

The model becomes overtrained very soon after 5 epochs for CGCL (Fig.6, left) and after 10 epochs for notMNIST (Fig.6, right) datasets. The prediction accuracy for test subset of 70 images and AOC for ROC-curve was very small (~0.5) and it is explained by the small size of datasets in comparison to the complexity of the CNN model.

To avoid such overtraining the lossless data augmentation with addition of the random horizontal and vertical flips of the original images was applied. In Fig.7 accuracy and loss results of training and validation attempts are shown for subsets of the original CGCL (Fig.7, left) and notMNIST (Fig.7, right) datasets that contained glyphs of A and H letters only. Again model became overtrained a little bit later: after 7 epochs for CGCL and after 15 epochs for notMNIST datasets. In this case the prediction accuracy for test subset and AOC for ROC-curve was very small (~0.5) also.

But application of the lossy data augmentation with addition of the random rotations (up to 40 degrees), width shifts (up to 20%), height shifts (up to 20%), shear (up to 20%), and zoom (up to 20%) allowed significantly increase both datasets and improve accuracy and loss without overtraining (Fig.8).

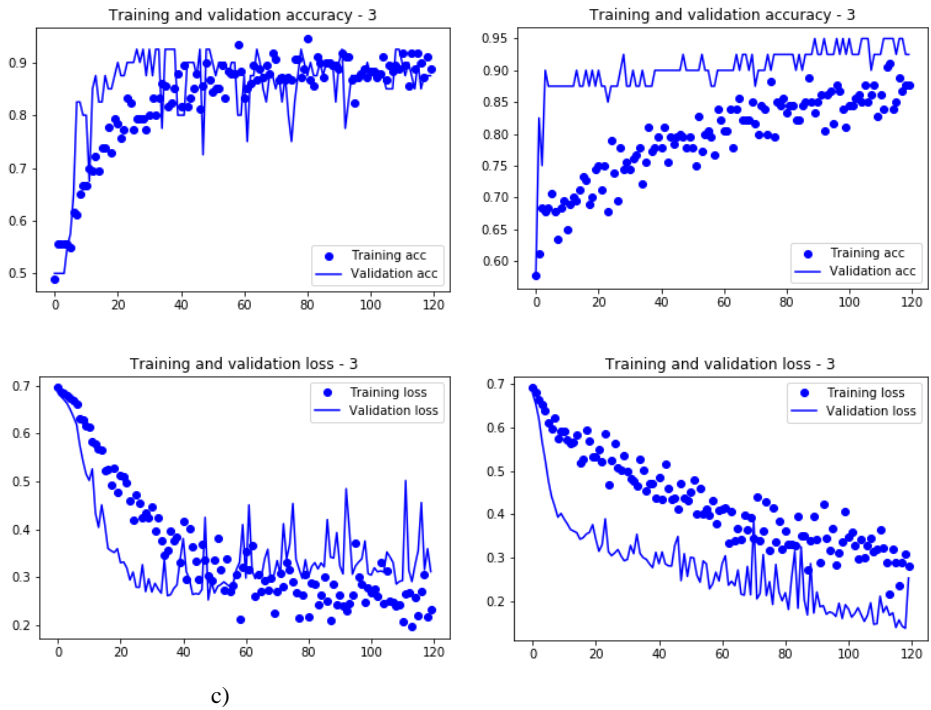

c)

**Fig. 8.** The accuracy and loss of training and validation attempts for the original CGCL (left) and notMNIST (right) datasets with lossy data augmentation.

As a result the prediction for the test subset became much better with accuracy 0.94, loss 0.21, and area AOC 0.99 for CGCL(Fig.9, left), and accuracy 0.91, loss 0.21, and area AOC 0.99 for ROC-curve for notMNIST (Fig.9, right) datasets.

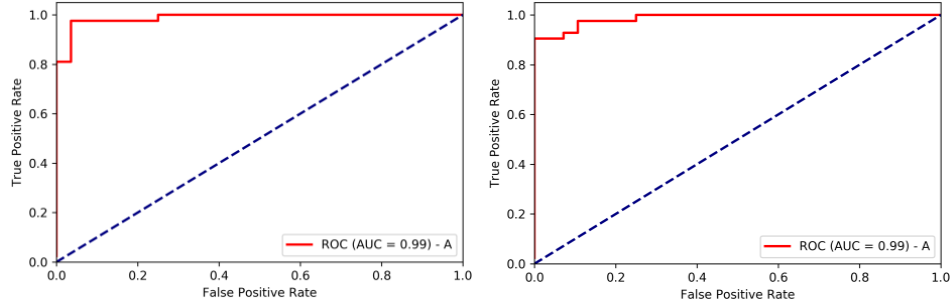

**Fig. 9.** ROC-curves for subsets of the original CGCL (a) and notMNIST (b) datasets that contained glyphs of A and H letters only.

## 6    Discussion and Future Work

The explanatory data analysis of CGCL and notMNIST datasets shown that the carved letters from CGCL can hardly be differentiated by dimensionality reduction methods, for example, by tSNE due to the worse letter representation by stone carving in comparison to glyphs of handwritten and printed letters like notMNIST. The results of MLR are good enough for the small dataset even, if the quality of glyphs is high enough, for example like in the cleaned part of notMNIST dataset. But for the more complicated glyphs like ones from CGCL dataset, MLR can provide quite mediocre predictions. In contrast, the CNN models gave the very high AUC values close to 0.99 for both notMNIST and CGCL (despite the much smaller size of CGCL in comparison to notMNIST) under condition of the high lossy data augmentation. That is why in the wider context the obtained models can be significantly improved to be very sensitive to many additional aspects like date, language, authorship, genuineness, and meaning of graffiti, for example, by usage of the capsule-based deep neural networks that were recently proposes and demonstrated on MNIST dataset [22]. The first attempts of application of such capsule-based deep neural networks seem to be very promising for classification problems in the context of notMNIST and CGL datasets [23]. But for this the much larger datasets and additional research of specifically tuned models will be necessary. In this context, the further progress can be reached by sharing the similar datasets around the world in the spirit of open science, volunteer data collection, processing and computing [2,21].

In conclusion, the new image dataset of the carved Glagolitic and Cyrillic letters was prepared and tested by MLR and deep CNNs for the letter recognition. The dataset was published for the data science community as an open source resource.

## Acknowledgements

The work was partially supported by Huizhou Science and Technology Bureau and Huizhou University (Huizhou, P.R.China) in the framework of Platform Construction for China-Ukraine Hi-Tech Park Project. The glyphs of letters from the graffiti [3] were prepared by students and teachers of National Technical University of Ukraine "Igor Sikorsky Kyiv Polytechnic Institute" and can be used as an open science dataset under CC BY-NC-SA 4.0 license (https://www.kaggle.com/yoctoman/graffiti-st-sophia-cathedral-kyiv).

## References

1. Ancelet, J.: The history of graffiti. University of Central London (2006).
2. Burt, D.: The Online Corpus of the Inscriptions from Ancient North Arabia (OCIANA), http://krc2.orient.ox.ac.uk/ociana, last accessed 2018/08/30.
3. Nikitenko, N., Kornienko, V.: Drevneishie Graffiti Sofiiskogosobora v Kieve i Vremya Ego Sozdaniya (Old Graffiti in the St. Sofia Cathedral in Kiev and Time of Its Creation), Mykhailo Hrushevsky Institute of Ukrainian Archeography and Source Studies, Kiev (in Russian) (2012).
4. Vysotskii, S. A.: Drevnerusskie Nadpisi Sofii Kievskoi XI—XIV vv. (Old Russian Inscriptions in the St. Sofia Cathedral in Kiev, 11th—14th Centuries) Kiev: Naukova Dumka (in Russian) (1966).
5. Nazarenko, T.: East Slavic Visual Writing: The Inception of tradition. Canadian Slavonic Papers, 43(2-3), 209-225 (2001).
6. Drobysheva, M.: The Difficulties of Reading and Interpretation of Old Rus Graffiti (the Inscription Vys. 1 as Example). Istoriya, 6(6 (39)), 10-20 (2015).
7. Pritsak, O.: An Eleventh-Century Turkic Bilingual (Turko-Slavic) Graffito from the St. Sophia Cathedral in Kiev. Harvard Ukrainian Studies, 6(2), 152-166 (1982).
8. LeCun, Y., Cortes, C., Burges, C. J.: MNIST Handwritten Digit Database. AT&T Labs. http://yann.lecun.com/exdb/mnist, last accessed 2018/08/30.
9. LeCun, Y., Bottou, L., Bengio, Y., and Haffner, P.: Gradient-based Learning Applied to Document Recognition. In Proceedings of the IEEE, 86(11), 2278–2324, IEEE (1998).
10. Hafemann, L. G., Sabourin, R., Oliveira, L. S.: Offline Handwritten Signature Verification — Literature Review. In 2017 Seventh International Conference on Image Processing Theory, Tools and Applications, pp. 1-8, IEEE (2017).
11. Winter, J.: Preliminary Investigations on Chinese Ink in Far Eastern Paintings, Archaeological Chemistry, 207-225 (1974).
12. Peng Gang, Jiang Hui, S. Stirenko, Yu. Gordienko, T. Shemsedinov, O. Alienin, Yu. Kochura, N. Gordienko, A. Rojbi, J.R. López Benito, E. Artetxe González, User-driven Intelligent Interface on the Basis of Multimodal Augmented Reality and Brain-Computer Interaction for People with Functional Disabilities, In Proc Future of Information and Communication Conference, 5-6 April 2018, pp.322-331, IEEE, Singapore (2018).
13. Gordienko, Yu., Stirenko, S., Alienin, O., Skala, K., Soyat, Z., Rojbi, A., López Benito, J.R., Artetxe González, E., Lushchyk, U., Sajn, L., Llorente Coto, A., Jervan G.: Augmented Coaching Ecosystem for Non-obtrusive Adaptive Personalized Elderly Care on the Basis of Cloud-Fog-Dew Computing Paradigm, In Proc. IEEE 40th International Conven-tion on Information and Communication Technology, Electronics and Microelectronics, pp.387-392, IEEE, Opatija, Croatia (2017).

14. Nikitenko, N., Kornienko, V.: Drevneishie Graffiti Sofiiskogosobora v Kieve i Ego Datirovka (The Ancient Graffiti of St. Sophia Cathedral in Kiev and Its Dating), Byzantinoslavica, 68(1), 205-240 (in Russian) (2010).

15. Kornienko, V.V.: Korpus Hrafiti Sofii Kyivskoi, XI - pochatok XVIII_st, chastyny I-III (The Collection of Graffiti of St. Sophia of Kyiv, 11th – 17th centuries), Parts I-III, MykhailoHrushevsky Institute of Ukrainian Archeography and Source Studies, Kiev (in Ukrainian), (2010-2011).

16. Sinkevič, N., Korniênko, V.: Nowe Źródła do Historii Kościoła unickiego w Kijowie: Graffiti w Absydzie Głównego Ołtarza Katedry Św. Zofii, Studia Źródłoznawcze, 50 (2012).

17. Glyphs of Graffiti in St. Sophia Cathedral of Kyiv, https://www.kaggle.com/yoctoman/graffiti-st-sophia-cathedral-kyiv, last accessed 2018/08/30.

18. Bulatov, Y.: notMNIST dataset. Google (Books/OCR), Tech. Rep., http://yaroslavvb.blogspot. it/2011/09/notmnist-dataset.html, last accessed 2018/08/30.

19. Maaten, L. V. D., Hinton, G.: Visualizing Data Using t-SNE. Journal of Machine Learning Research, 9(Nov), 2579-2605 (2008).

20. Schmidt, P.: Cervix EDA and model selection, https://www.kaggle.com/philschmidt, last accessed 2018/08/30.

21. Gordienko, N., Lodygensky, O., Fedak, G., Gordienko, Yu.: Synergy of volunteer measurements and volunteer computing for effective data collecting, processing, simulating and analyzing on a worldwide scale, In: Proc. 38th International Convention on Information and Communication Technology, Electronics and Microelectronics, pp. 193-198, IEEE, Opatija, Croatia (2015).

22. Sabour, S., Frosst, N., Hinton, G. E.: Dynamic Routing Between Capsules, In Advances in Neural Information Processing Systems, pp. 3856-3866 (2017).

23. Gordienko, N., Kochura, Yu., Taran, V., Gang, P., Gordienko, Yu., Stirenko, S.: Capsule Deep Neural Network for Recognition of Historical Graffiti Handwriting, IEEE Ukraine Student, Young Professional and Women in Engineering Congress, October 2-6, 2018, Kyiv, Ukraine, IEEE (2018) (submitted).